\documentclass[10pt,twocolumn,letterpaper]{article}

\usepackage[pagenumbers]{cvpr} 

\usepackage{multirow}
\usepackage{pifont}
\usepackage{arydshln}
\usepackage[accsupp]{axessibility}  
\usepackage{balance}

\definecolor{cvprblue}{rgb}{0.21,0.49,0.74}
\usepackage[pagebackref,breaklinks,colorlinks,allcolors=cvprblue]{hyperref}
\usepackage{orcidlink}

\def\paperID{*****} 
\def\confName{CVPR}
\def\confYear{2025}

\title{Dual-Path Enhancements in Event-Based Eye Tracking: Augmented Robustness and Adaptive Temporal Modeling}

\author{Hoang M. Truong\orcidlink{0009-0008-3899-895X}\and
Vinh-Thuan Ly\orcidlink{0009-0005-7557-8231}\and
Huy G. Tran\orcidlink{0009-0001-2580-2355}\and
Thuan-Phat Nguyen\orcidlink{0009-0005-2152-1336}\and
Tram T. Doan\orcidlink{0000-0003-3352-8082}\\
University of Science, VNU-HCM, Ho Chi Minh City, Vietnam\\
Vietnam National University, Ho Chi Minh City, Vietnam\\
{\tt\small \{22280034, 22280092, 22280040, 22280062\}@student.hcmus.edu.vn, dttram@hcmus.edu.vn}
}

\begin{document}
\maketitle
\begin{abstract}
Event-based eye tracking has become a pivotal technology for augmented reality and human-computer interaction. Yet, existing methods struggle with real-world challenges such as abrupt eye movements and environmental noise. Building on the efficiency of the Lightweight Spatiotemporal Network—a causal architecture optimized for edge devices—we introduce two key advancements. First, a robust data augmentation pipeline incorporating temporal shift, spatial flip, and event deletion improves model resilience, reducing Euclidean distance error by 12\% (1.61 vs. 1.70 baseline) on challenging samples. Second, we propose KnightPupil, a hybrid architecture combining an EfficientNet-B3 backbone for spatial feature extraction, a bidirectional GRU for contextual temporal modeling, and a Linear Time-Varying State-Space Module to adapt to sparse inputs and noise dynamically. Evaluated on the 3ET+ benchmark, our framework achieved 1.61 Euclidean distance on the private test set of the Event-based Eye Tracking Challenge at CVPR 2025, demonstrating its effectiveness for practical deployment in AR/VR systems while providing a foundation for future innovations in neuromorphic vision.
\end{abstract}    
\section{Introduction}
\label{sec:intro}

The human eye's remarkable agility - capable of executing saccades at speeds exceeding 500°/s~\cite{ROLFS20092415} - presents both an inspiration and a challenge for machine vision systems. As augmented reality headsets evolve toward gaze-contingent rendering and neurological diagnostics increasingly rely on precise oculomotor analysis~\cite{10.1145/3448018.3458016}, the limitations of conventional frame-based eye tracking become apparent: their temporal resolution and power efficiency fundamentally constrain applications demanding real-time responsiveness. Event cameras, with their bio-inspired asynchronous operation~\cite{9138762}, offer a paradigm shift. These sensors achieve microsecond temporal resolution while consuming merely 10mW - two orders of magnitude more efficient than high-speed cameras~\cite{Maqueda_2018}. However, their sparse, irregular output streams defy traditional computer vision approaches. The core challenge lies in developing algorithms that can extract robust gaze estimates from these noisy, intermittent signals while maintaining real-time performance on edge devices~\cite{6887319}.

Recent research has significantly advanced event-based eye tracking. Ding et al.~\cite{ding2024facet} introduced FACET, which uses fast ellipse fitting to enhance gaze estimation accuracy, particularly for XR applications. However, it may face challenges under occlusion or varying lighting conditions. Similarly, Chen et al.~\cite{chen20233et} proposed 3ET, utilizing a Change-based ConvLSTM network to preserve temporal precision from raw event data. Despite strong performance, the recurrent nature of ConvLSTM can introduce latency issues on resource-constrained edge devices. Additionally, the Lightweight Spatiotemporal Network~\cite{pei2024lightweightspatiotemporalnetworkonline} demonstrated effectiveness through its causal architecture and efficient FIFO-based processing, though practical deployment highlights challenges such as blink-induced artifacts and interference from eyewear. Overcoming these limitations holds great potential for consumer AR/VR, enabling always-on gaze interaction without draining battery life, and for clinical settings, where it could aid in monitoring digital biomarkers for neurodegenerative diseases~\cite{10.1145/3448018.3458016}.

Current event-based eye tracking methods typically follow one of two paths: either compressing events into frame-like representations~\cite{rebecq2019highspeedhighdynamic} or processing raw events through specialized architectures~\cite{pei2024lightweightspatiotemporalnetworkonline}. While the latter approach preserves temporal precision, most implementations struggle to balance computational efficiency with adaptability to challenging real-world conditions. Recent literature reflects a growing interest in hybrid approaches. Some works combine convolutional and recurrent networks, while others explore novel spiking neural architectures~\cite{yin2020effectiveefficientcomputationmultipletimescale}. However, these often introduce substantial computational overhead or require complex training protocols, limiting their practical applicability. 

The Lightweight Spatiotemporal Network~\cite{pei2024lightweightspatiotemporalnetworkonline} offers efficient event processing via a causal design, FIFO buffering, and depthwise convolutions, making it well-suited for constrained environments. However, our experiments show that its fixed temporal receptive field limits performance under variable eye movements and real-world noise.

We explore two complementary solutions. First, we improve the original network using data augmentation techniques such as temporal shifting and spatial flipping, enhancing its robustness without compromising efficiency. Second, we propose KnightPupil, a novel architecture combining EfficientNet-based spatial encoding, bidirectional GRUs, and an LTV-SSM for adaptive temporal modeling. While distinct, both strategies contribute to advancing real-time, robust event-based gaze estimation by balancing efficiency and adaptability.

Building upon recent advances in event-based vision, we make three key contributions:
\begin{itemize}
    \item Augmented Robustness for Spatiotemporal Networks: Strategic data augmentation (temporal shift, spatial flip, event deletion) improves resilience to real-world perturbations while preserving computational efficiency.
    \item Adaptive Hybrid Architecture: KnightPupil integrates spatial feature extraction (EfficientNet-B3), bidirectional temporal modeling (GRU), and dynamic state adaptation (LTV-SSM) for noise-resistant gaze estimation.
    \item Dual-Strategy Framework: Combines a deployable optimized baseline with an exploratory architecture, advancing both practical deployment and adaptive temporal modeling research.
\end{itemize}
\section{Related Work}
\label{sec:related_work}

\subsection{Lightweight Spatiotemporal Network}
Event-based vision has gained significant attention due to its high temporal resolution and sparse nature, making it suitable for real-time applications such as eye tracking~\cite{Gallego_2022}. Traditional approaches often rely on frame-based processing, which can lose valuable temporal information when applied to event-based data. To address this, spatiotemporal neural networks have been developed to leverage both spatial and temporal event patterns.

\begin{figure}[ht]
    \centering
    \includegraphics[width=1\linewidth]{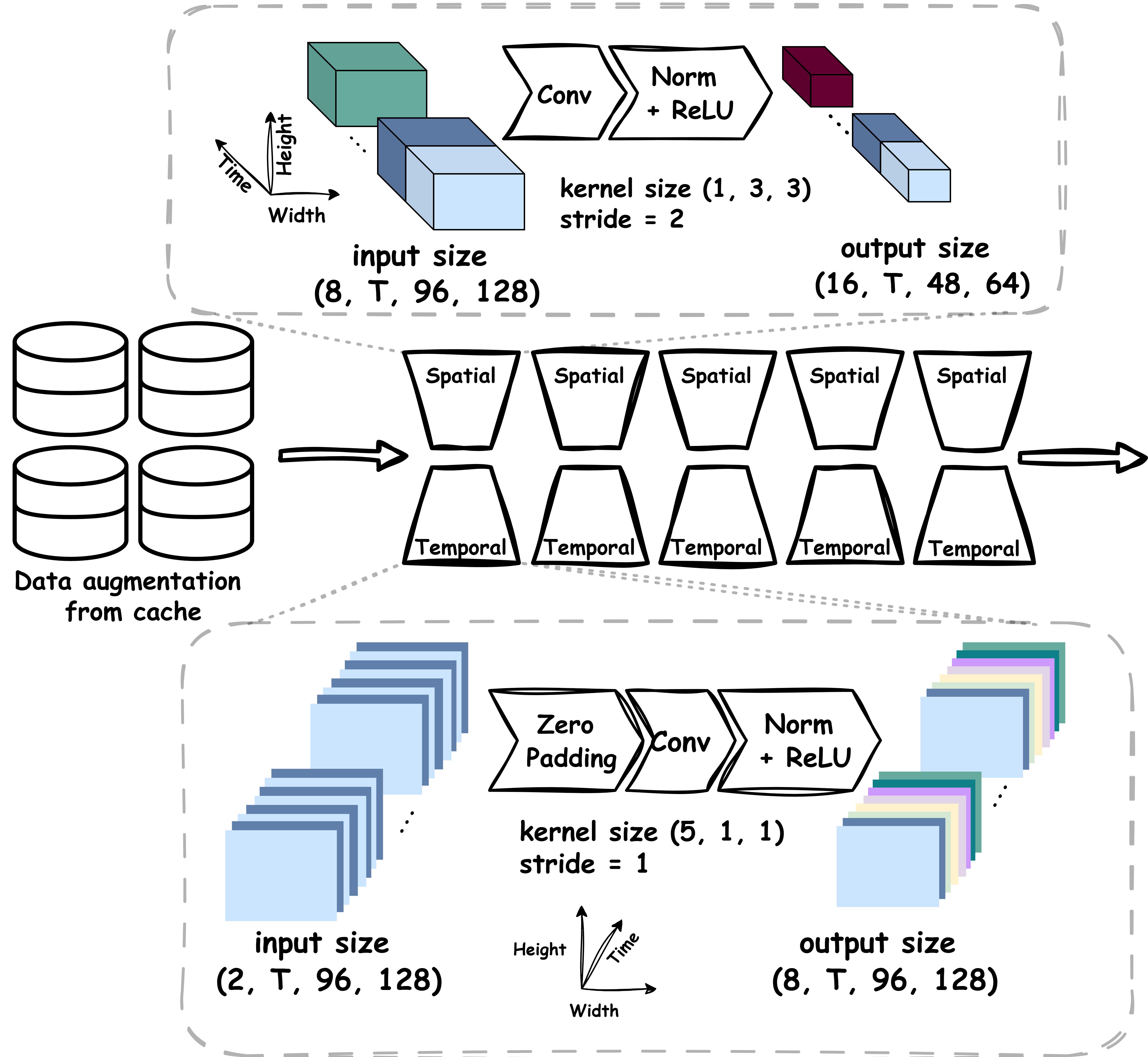}
    \caption{A compact spatiotemporal model integrating data augmentation with spatial and temporal processing blocks. Convolutional layers extract spatial and temporal features efficiently.}
    \label{fig:spatiotemporal_architecture}

    \vspace{0.3cm}

    \includegraphics[width=1\linewidth]{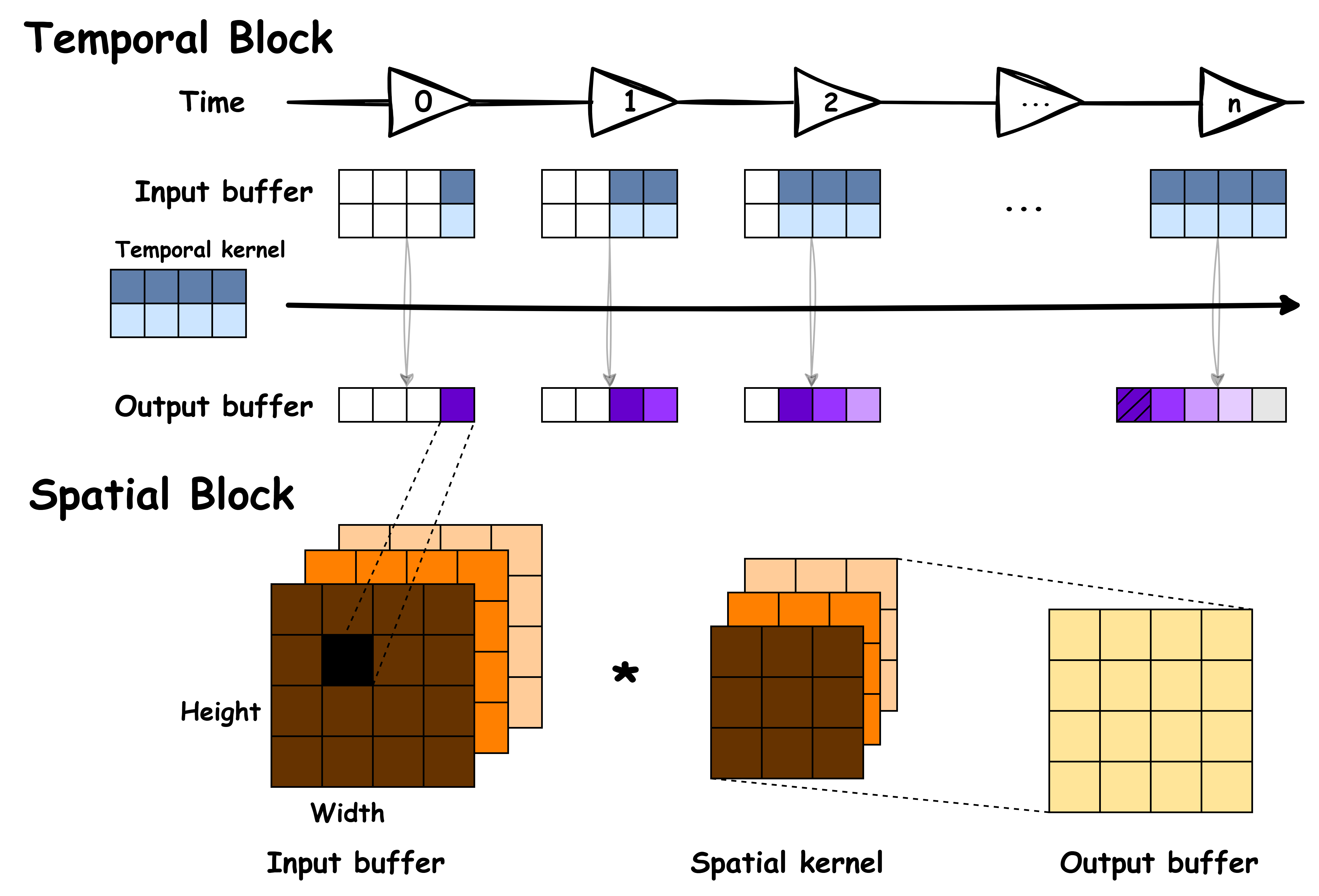}
    \caption{Illustration of spatiotemporal processing: the Temporal Block applies temporal convolution across frames, while the Spatial Block extracts spatial features using convolutional filters.}
    \label{fig:spatiotemporal_inference}
\end{figure}

The lightweight spatiotemporal network proposed in this domain is a notable advancement proposed in \cite{pei2024lightweightspatiotemporalnetworkonline}. This network employs a causal spatiotemporal convolutional architecture designed to perform online inference efficiently. The key components include:

\begin{itemize}
    \item \textbf{Causal Spatiotemporal Convolutional Blocks:} The network utilizes a sequence of spatiotemporal blocks, each consisting of a temporal convolution followed by a spatial convolution. This structure ensures efficient temporal feature extraction while maintaining causality, allowing real-time inference.
    \item \textbf{Causal Event Binning:} Unlike conventional binning techniques that introduce latency, this method causally processes events, reducing buffering requirements and improving efficiency in streaming data applications.
    \item \textbf{Hybrid Normalization Strategy:} The network alternates between Batch Normalization (BN) and Group Normalization (GN) to balance training stability and adaptability across different batch sizes, enhancing generalization.
    \item \textbf{Sparse Activation Regularization:} An L1 regularization term is applied to activations to encourage sparsity, significantly reducing computational complexity while maintaining accuracy.
\end{itemize}

Experimental results demonstrate that this lightweight architecture achieves state-of-the-art performance on event-based eye-tracking tasks, particularly in terms of p10 accuracy and computational efficiency. Integrating efficient event binning, causal convolutions, and activation sparsity techniques makes it well-suited for deployment on edge devices.

Our work builds upon this foundation by incorporating an advanced data augmentation strategy, further enhancing model robustness and generalization. The lightweight spatiotemporal architecture and its inference pipeline are shown in~\cref{fig:spatiotemporal_architecture} and~\cref{fig:spatiotemporal_inference}.

\subsection{Voxel Grid Representation in Event-based Processing}  

Voxel grids have been widely adopted in event-based vision for spatiotemporal feature encoding. Unlike frame-based methods that rely on regularly sampled intensity values, voxel grids aggregate asynchronous events into structured 3D tensors, where the temporal dimension is discretized into multiple bins. This representation effectively bridges the gap between event-based sensing and deep learning, enabling efficient feature extraction while preserving fine-grained motion information.  

Early works such as \cite{clady2017voxel} explored voxel-based accumulation to enhance event representation for motion estimation. Subsequent research extended this approach, integrating voxel grids into convolutional neural networks for tasks like object recognition \cite{gehrig2019end} and optical flow estimation \cite{zhu2019unsupervised}. More recently, voxel-based encoding has been utilized in transformer-based models for event-based classification, further demonstrating its versatility \cite{stoffregen2020reducing}.  

Despite their effectiveness, voxel grids introduce a trade-off between temporal resolution and memory efficiency. Selecting an appropriate number of bins is crucial: too few bins may discard valuable temporal details, while too many can increase computational overhead. Our approach employs a fixed 3-bin voxel grid, ensuring compatibility with EfficientNet’s input structure while maintaining a balance between temporal precision and computational efficiency.


\subsection{EfficientNet}  
EfficientNet~\cite{tan2019efficientnet} is a family of CNNs designed to optimize accuracy and efficiency through compound scaling. It achieves state-of-the-art performance with significantly fewer parameters compared to traditional deep networks. Among its variants, EfficientNet-B3 provides a strong balance between model complexity and inference speed, making it well-suited for vision tasks requiring real-time processing.  

In this work, we leverage EfficientNet-B3 as a feature extractor for event-based eye tracking, providing a compact yet expressive representation of input data. By integrating EfficientNet-B3 with a sequential model, we aim to capture both spatial and temporal dependencies, enabling robust and efficient gaze estimation.

\subsection{State-Space Models for Sequential Modeling}  
A \textbf{State-Space Model (SSM)} is a mathematical framework that models dynamic systems using latent states that evolve over time. SSMs have been widely explored in sequence modeling due to their ability to capture long-range dependencies~\cite{gu2022efficiently, gu2023mamba} efficiently. Compared to recurrent neural networks (RNNs), SSMs offer structured representations of temporal dynamics, making them particularly useful in applications where memory efficiency and stability are crucial.

Recent works have extended SSMs to event-based vision tasks, demonstrating their potential for modeling sparse and asynchronous data streams~\cite{zubic2024state}. However, conventional SSMs often involve complex parameterization and high computational cost, making them less practical for real-time applications.  

To address this, we introduce a \textbf{simplified version of the Linear Time-Variant State-Space Model (LTV-SSM)}, specifically designed to complement the \textbf{EfficientNet-B3 + GRU} architecture. By keeping the formulation lightweight while preserving key temporal modeling capabilities, our approach maintains computational efficiency without sacrificing predictive power. This enables our model to effectively capture spatial and temporal dependencies in event-based eye tracking, improving robustness and accuracy.
\section{Methodology}
\label{sec:methodology}

Our approach combines data augmentation and a lightweight neural architecture to enhance event-based eye-tracking. We first apply augmentation techniques to improve model robustness to temporal and spatial variations. Then, we introduce KnightPupil, which integrates voxel grid encoding, EfficientNet-B3 for spatial features, Bi-GRU for temporal modeling, and LTV-SSM for adaptive state transitions, enabling precise and efficient gaze estimation.

\subsection{Data Augmentation}

\begin{figure}[t]
  \centering
  \includegraphics[width=1\linewidth]{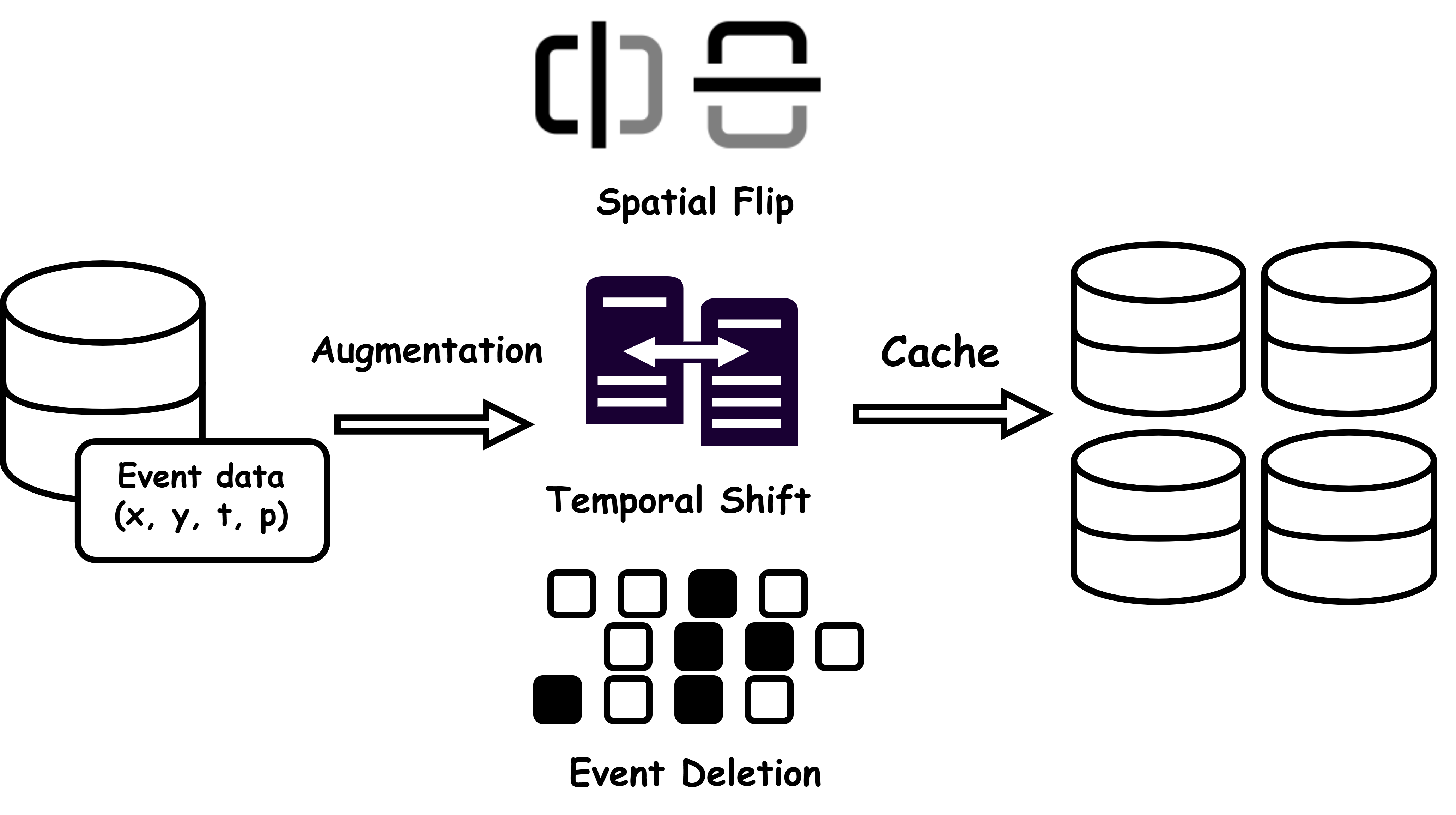}
  \caption{Overview of data augmentation techniques: Spatial Flip mirrors event coordinates, Temporal Shift modifies event timing, and Event Deletion simulates sensor noise.}
  \label{fig:data_augmentation}
\end{figure}

To enhance the robustness and generalizability of our event-based eye-tracking model, we employ a series of augmentation techniques tailored for event streams. These augmentations include \textbf{temporal shift}, \textbf{spatial flip}, and \textbf{event deletion}, each designed to introduce realistic variations while preserving label consistency. These augmentation techniques are visually overviewed in~\cref{fig:data_augmentation}.

\paragraph{Temporal Shift.} 
Given the asynchronous nature of event-based data, temporal augmentation is crucial for improving model resilience to timing variations. We apply a random shift $\Delta t$ to event timestamps $t_i$ within a range of $\pm200$ milliseconds:
\begin{equation}
    t_i' = t_i + \Delta t, \quad \Delta t \sim \mathcal{U}(-200, 200) \text{ ms},
\end{equation}
where $\mathcal{U}(a, b)$ denotes a uniform distribution. Since labels are sampled at 100Hz (every 10ms), we recompute the label indices as
\begin{equation}
    L_j' = L_{j + \lfloor \Delta t / 10 \rfloor},
\end{equation}
ensuring accurate correspondence between events and labels.

\paragraph{Spatial Flip.} 
To introduce spatial invariance, we horizontally and vertically flip the event coordinates $(x, y)$:
\begin{equation}
    x' = W - x, \quad y' = H - y,
\end{equation}
where $W$ and $H$ are the width and height of the sensor, respectively. The corresponding pupil center labels undergo the same transformation:
\begin{equation}
    (x_{\text{label}}', y_{\text{label}}') = (W - x_{\text{label}}, H - y_{\text{label}}).
\end{equation}

\paragraph{Event Deletion.} 
To simulate real-world sensor noise and occlusions, we randomly remove $p=5\%$ of the events:
\begin{equation}
    P(\text{delete } e_i) = p, \quad e_i \in E,
\end{equation}
where $E$ is the set of all events. The label sequence remains unchanged, ensuring the model learns to handle missing data effectively.

Each augmentation is applied independently to the training set, generating multiple augmented versions of the original data. The final dataset comprises both the original and augmented samples, ensuring diversity in both spatial and temporal domains while maintaining label accuracy.

\subsection{KnightPupil} 

\begin{figure}[tp]
  \centering
  \includegraphics[width=1\linewidth]{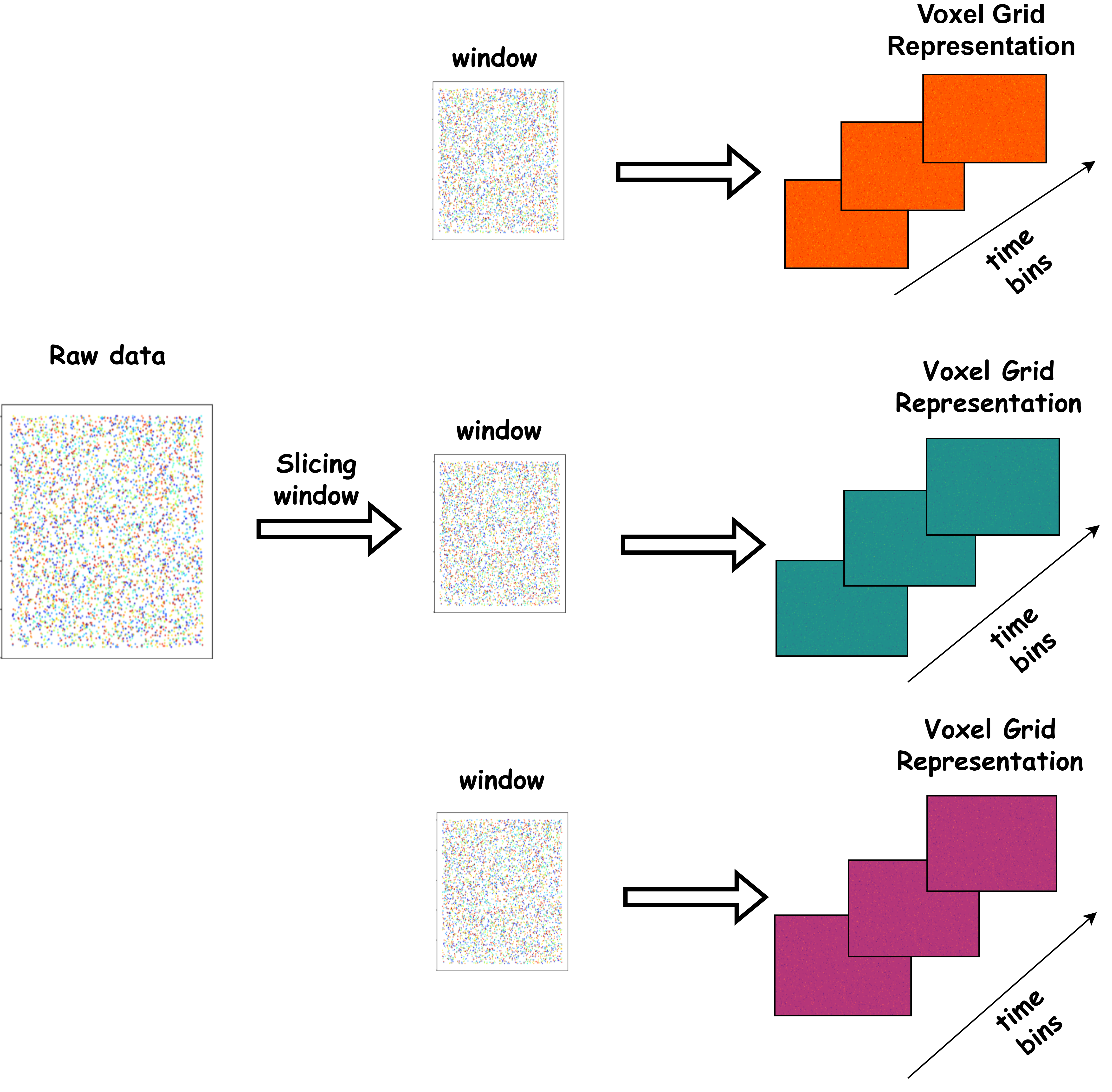}
  \caption{Voxel grid representation of event data, where raw events are segmented into windows and accumulated into spatial bins over time.}
  \label{fig:voxel_grid}
\end{figure}

Our proposed eye-tracking solution integrates EfficientNet-B3 for spatial feature extraction, Bi-GRU with dropout for temporal modeling, and a linear time-varying state-space model (LTV-SSM) for adaptive state transitions. This architecture is designed to effectively process event-based data by efficiently capturing both spatial and temporal dependencies.

\begin{figure*}[htbp]
  \centering
  \includegraphics[width=1\linewidth]{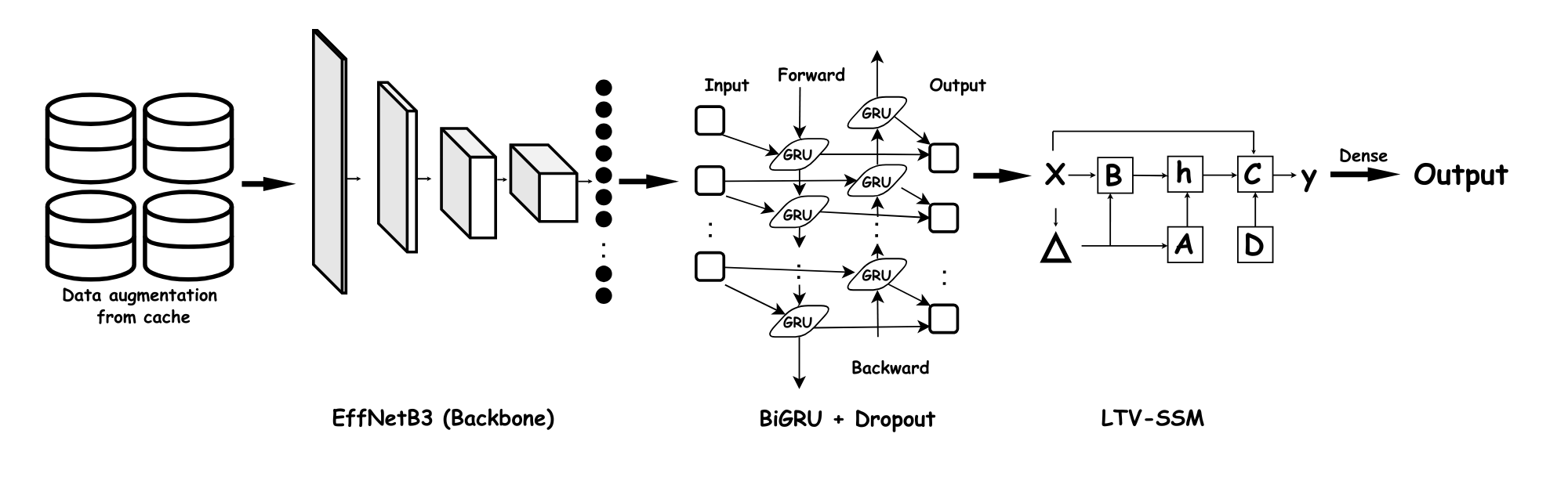}
  \caption{Overview of the KnightPupil architecture. The model consists of three key components: (1) an EfficientNet-B3 backbone for spatial feature extraction, (2) a Bidirectional GRU (Bi-GRU) for temporal modeling, and (3) a Linear Time-Varying State-Space Model (LTV-SSM) for adaptive state transitions. This design enables robust event-based gaze estimation by efficiently capturing spatial and temporal dependencies.}
  \label{fig:KnightPupil}
\end{figure*}

We name our model KnightPupil to reflect its strengths: “Knight” symbolizes robustness and adaptability, highlighting its resilience with dynamic event-based data. "Pupil" signifies its focus on eye-tracking, drawing a direct connection to its primary task. KnightPupil embodies a vigilant and efficient system for gaze estimation, capable of precisely tracking rapid eye movements. The architecture is illustrated in~\cref{fig:KnightPupil}.

\paragraph{Affine Transformation and KnightPupil}  
For KnightPupil, we chose not to use an affine transformation in the preprocessing stage, unlike the augmentation technique used in the model proposed by Pei et al.~\cite{pei2024lightweightspatiotemporalnetworkonline}. Instead, our approach leverages raw event data directly for augmentation. This method allows us to maintain the integrity of temporal and spatial details crucial for event-based eye-tracking, ensuring efficient training while avoiding unnecessary transformations that could distort these key characteristics in KnightPupil’s architecture.

\paragraph{Voxel Grid Representation}  

Since event cameras produce a stream of asynchronous events rather than conventional frames, we employ voxel grid encoding to transform these events into a structured representation. Given a set of events \( \mathcal{E} = \{(x_i, y_i, t_i, p_i)\}_{i=1}^{N} \), where \( (x_i, y_i) \) represents the spatial coordinates, \( t_i \) is the timestamp, and \( p_i \in \{-1, 1\} \) denotes the polarity, we construct a voxel grid representation \( V \in \mathbb{R}^{H \times W \times T} \) by discretizing the time domain into \( T \) bins and accumulating events within each bin. The illustration of the voxel grid is shown in~\cref{fig:voxel_grid}.

Following the implementation in \texttt{Tonic}, the voxel grid is computed as:
\begin{equation}
V(x, y, t) = \sum_{i=1}^{N} p_i \cdot \max \left(0, 1 - \left| t - \frac{T (t_i - t_{\text{min}})}{t_{\text{max}} - t_{\text{min}}} \right| \right),
\end{equation}
where the temporal bin index is computed using a linear interpolation scheme to distribute events across adjacent bins. This ensures a smooth temporal representation while preserving event polarity information. The voxel grid is normalized such that each bin contains values scaled between -1 and 1, improving numerical stability for training.

This transformation enables \textbf{KnightPupil} to efficiently process event-based eye-tracking data, capturing both spatial and temporal event distributions with enhanced resolution and robustness.

\paragraph{Spatial Feature Extraction.}  
We use EfficientNet-B3 as the backbone to extract spatial features in our event-based eye-tracking framework. Its optimized architecture balances efficiency and representational power, making it ideal for capturing spatial patterns. EfficientNet scales depth, width, and resolution using a compound coefficient \( \phi \):
\begin{align}
    d &= \alpha^\phi d_0, \\
    w &= \beta^\phi w_0, \\
    r &= \gamma^\phi r_0,  
\end{align}
where \( d \), \( w \), and \( r \) represent the depth, width, and input resolution, respectively. The constants \( \alpha, \beta, \gamma \) are determined via grid search, and \( \phi \) controls the overall scaling. For EfficientNet-B3, the specific values are \( \phi = 1.8 \), \( \alpha = 1.2 \), \( \beta = 1.1 \), and \( \gamma = 1.15 \), resulting in a deeper network with more channels and higher resolution compared to EfficientNet-B0.  

We initialize EfficientNet-B3 with pre-trained ImageNet weights to process event-based eye-tracking data. Unlike partial fine-tuning approaches that freeze lower layers, we perform full fine-tuning, allowing all layers to be updated during training. This ensures that the model fully adapts to the event-based eye-tracking task rather than relying solely on generic ImageNet features.  
Given an input voxel grid representation \( V \), the extracted feature representation is:  
\begin{equation}
    F = \text{EffNetB3}(V)
\end{equation}
where \( F \in \mathbb{R}^{T \times d} \) is the sequence of extracted spatial features.  

We chose EfficientNet-B3 over other commonly used convolutional networks such as ResNet for several reasons. First, EfficientNet employs compound scaling, allowing it to achieve a superior trade-off between accuracy and efficiency compared to traditional architectures. In contrast, ResNet scales depth independently, which may lead to redundancy in feature extraction without proportionally improving performance. Additionally, EfficientNet's depthwise-separable convolutions significantly reduce computational cost, making it more suitable for real-time applications like eye tracking.  

Another crucial factor is EfficientNet's superior parameter efficiency. ResNet architectures, particularly deeper variants such as ResNet-50 or ResNet-101, contain many parameters, increasing both memory footprint and inference latency. In contrast, EfficientNet-B3 achieves comparable or better accuracy with fewer parameters, enabling more efficient deployment in resource-constrained environments. Moreover, EfficientNet has demonstrated strong generalization on various vision tasks, making it a robust choice for event-based processing where spatial features must be efficiently captured.  

By leveraging EfficientNet-B3 as our feature extractor, we ensure an optimal balance between accuracy, computational efficiency, and real-time applicability, making it well-suited for event-based eye tracking.

\paragraph{Temporal Modeling}  
The extracted feature sequence \( F \) is obtained from EfficientNet-B3, where each frame is encoded into a \( d \)-dimensional feature vector. In our case, \( d = 1536 \) for EfficientNet-B3. This sequence is then passed through a bidirectional GRU with two layers and a hidden size of 128 per direction, resulting in a total hidden size of 256. Dropout of 0.3 is applied between GRU layers to improve generalization. The GRU update equations are:  
\begin{align}  
    r_t &= \sigma(W_r F_t + U_r h_{t-1} + b_r), \\  
    z_t &= \sigma(W_z F_t + U_z h_{t-1} + b_z), \\  
    \tilde{h}_t &= \tanh(W_h F_t + U_h (r_t \odot h_{t-1}) + b_h), \\  
    h_t &= (1 - z_t) \odot h_{t-1} + z_t \odot \tilde{h}_t.  
\end{align}  
where \( r_t \) and \( z_t \) are the reset and update gates, and \( h_t \) represents the hidden state. Since we employ bidirectional GRU, the final output is concatenated from both forward and backward directions. The GRU outputs a feature sequence of shape \( (T, 256) \), which is then fed into LTV-SSM for further sequential modeling.  

For several reasons, we adopt a GRU–LTV-SSM combination over other temporal models like LSTMs and Transformers. While LSTMs are widely used, their additional gating mechanisms and cell states increase computational complexity and inference latency—critical factors in real-time applications like event-based eye tracking. GRUs offer comparable performance with reduced overhead, making them a better fit for our lightweight setup.

Despite their strength in modeling long-range dependencies, transformers require self-attention computations that scale quadratically with sequence length and lack an intrinsic temporal bias, relying instead on positional encodings. These limitations make them less suitable for long, high-frequency event sequences. In contrast, our hybrid model balances efficiency and expressiveness: GRUs handle short-term dependencies, while the LTV-SSM component captures dynamic state transitions, enabling robust and efficient gaze estimation.

\paragraph{Adaptive State-Space Modeling.}  
We employ a Linear Time-Varying State-Space Model (LTV-SSM) to enhance the sequential representation. Unlike traditional state-space models with fixed transition matrices, our approach dynamically learns state transitions based on the GRU output $ h_t $. Specifically, we parameterize the transition matrices as:  
\begin{align}
    \Delta A_t &= \exp(\delta_t), \\
    \Delta B_t &= \delta_t \odot B_t, \\
    h'_t &= \Delta A_t h_t + \Delta B_t.
\end{align}
where $ \delta_t $ and $ B_t $ are dynamically learned from the input $ h_t $ using a linear transformation. Here, $ \Delta A_t $ ensures numerical stability, while $ \Delta B_t $ refines the hidden state representation.  

Additionally, we maintain an identity matrix $ \mathbf{D} $ to stabilize transformations, resulting in the final output:  
\begin{equation}
    y_t = C h'_t + h_t \mathbf{D}^T.
\end{equation}
\paragraph{Comparison of LTV-SSM with Mamba and Gated Linear Attention (GLA).}
LTV-SSM shares conceptual similarities with recent state-space sequence models such as Mamba~\cite{gu2023mamba} and GLA~\cite{yang2023gated}, particularly in its ability to selectively process input sequences. Similar to Mamba, LTV-SSM leverages learned transformations to adjust state transitions based on input features dynamically. However, unlike Mamba, which employs selective long-range convolutions and gating mechanisms, LTV-SSM maintains a structurally simpler formulation with element-wise multiplicative updates, making it computationally lightweight while still capturing long-range dependencies. Compared to GLA, which integrates explicit gating into sequence modeling, LTV-SSM avoids additional gate parameters, reducing the overall model complexity while preserving adaptability. 
Furthermore, GLA is explicitly designed for hardware-efficient training by leveraging structured gating mechanisms to optimize memory and computational efficiency. In contrast, LTV-SSM prioritizes a lightweight design with fewer learnable parameters, making it more suitable for scenarios where model simplicity and efficiency are critical.

With the inclusion of BiGRU in the architecture, the use of LTV-SSM becomes even more suitable, as it complements the bidirectional processing of temporal dependencies already captured by BiGRU. The LTV-SSM's simpler, computationally efficient formulation avoids the need for a more complex state-space model, as it effectively models long-range dependencies through element-wise multiplicative updates. This makes LTV-SSM a more lightweight and appropriate choice in combination with BiGRU, eliminating the need for additional gating mechanisms or state-space models, while maintaining adaptability and reducing model complexity.
The LTV-SSM training process does not rely on parallel scan techniques. The model is trained using standard sequential operations in the typical deep learning training loop, where sequences are processed in a forward pass and gradients are computed via a backward pass. Parallel scan is not used, as the element-wise updates in the LTV-SSM allow for efficient computation without the need for complex parallelization strategies.

\paragraph{Final Prediction Layer.}  
The final gaze coordinates \((x, y)\) are obtained by applying a fully connected layer over the transformed sequence of hidden states from LTV-SSM:  
\begin{equation}
Y = W_o H' + b_o
\end{equation}
where \( H' \in \mathbb{R}^{T \times d} \) is the sequence of hidden states output by LTV-SSM, and \( W_o \) maps these features to a 2D coordinate space. The dropout layer before \( W_o \) helps regularize predictions by preventing overfitting.

By integrating EfficientNet-B3 for spatial feature extraction, Bi-GRU with dropout for temporal modeling, and LTV-SSM for adaptive state transitions, our solution maximizes both the benefits of pre-training and the adaptability of fine-tuning, leading to a more robust and high-performing eye-tracking model.

\paragraph{Voxel Grid Caching.}  
To improve training efficiency, we employ a disk caching mechanism using the \texttt{DiskCachedDataset} utility from the Tonic library. This approach minimizes redundant computations in subsequent epochs and future training sessions, significantly reducing data preprocessing overhead. While the initial transformation incurs some latency, caching allows the model to bypass repeated voxel grid construction, leading to much faster data loading in later iterations.  

Tonic’s caching mechanism stores preprocessed voxel grid representations on disk instead of RAM, ensuring efficient retrieval during training. This particularly benefits large-scale event-based datasets, where on-the-fly transformations can be computationally expensive. Furthermore, the caching system is designed to automatically detect changes in data transformations (such as modifications in bin size or augmentations), preventing outdated cached files from being reused incorrectly.  

Our approach accelerates training convergence by eliminating the need for redundant computations. It also facilitates experimentation with different hyperparameters or model configurations without excessive preprocessing delays. The cached dataset integrates seamlessly into the training pipeline, allowing for efficient and reproducible data loading. This optimization effectively balances preprocessing time and computational efficiency, making it well-suited for large-scale event-based eye-tracking applications.

\section{Experiment}
\label{sec:experiment}

\subsection{Datasets}

We conduct our experiments on the 2024 and 2025 3ET+ datasets~\cite{chen2025eventvision_event, wang2024ais_event}, both comprising 13 subjects, each with 2--6 recording sessions. The datasets share an event resolution of $640 \times 480$ and contain five activity classes: random, saccades, reading, smooth pursuit, and blinks.  

Ground truth annotations are provided at 100Hz, including pupil center coordinates $(x, y)$ and a binary blink indicator (\texttt{close}), where $0$ denotes an open eye and $1$ a closed eye. The evaluation is based on the average Euclidean distance between the predicted and ground truth pupil center coordinates. For the 2024 dataset, the percentage of cases where the average Euclidean distance is less than 10 (p10) is used.

\subsection{Implementation Details}

We conduct all experiments on a single NVIDIA Tesla P100 GPU provided by Kaggle.

For the lightweight spatiotemporal network, we adopt the original training configuration from Pei \etal~\cite{pei2024lightweightspatiotemporalnetworkonline}. Specifically, we train the model using a batch size of 32, each containing 50 event frames. The network is optimized for 200 epochs using the AdamW optimizer with a base learning rate of 0.002 and a weight decay of 0.005. We employ a cosine decay learning rate schedule with linear warmup, where the warmup phase spans 2.5\% of the total training steps. Additionally, we leverage automatic mixed-precision (AMP, FP16) and PyTorch compilation to accelerate training and improve efficiency.

For KnightPupil, we adopt a structured training approach to optimize its performance on event-based eye-tracking. The model is trained for 600 epochs with a batch size of 24, using the Adam optimizer with an initial learning rate of 0.001. A StepLR scheduler reduces the learning rate by 0.5 every 200 epochs. To handle event-based input efficiently, we convert raw event streams into voxel grids with num time bins = 3, where each sample spans a 0.3s time window (train length = 30 and temporal subsample factor = 1). Events within this window are binned into three equal sub-intervals of 0.1s each, forming an 80×60×3 voxel grid due to the spatial downsampling factor of 0.125. We employ a sliding window strategy, where training samples have a stride of 0.15s (train stride = 15), ensuring overlap, while validation samples use a 0.3s stride (val stride = 30) for non-overlapping evaluation. The model is validated at regular intervals, and we save the best checkpoint based on both the lowest Euclidean error and validation loss, along with the final model at the last epoch.

\subsection{Results and Ablation Study}

\paragraph{Effect of Augmentation.}
We evaluate the impact of our augmentation techniques on the architectures integrated in this study. The results are presented in~\cref{tab:euclidean_comparison}.

\begin{table}[ht]
    \centering
    \footnotesize
    \begin{tabular}{clccc}
        \toprule
        \textbf{Year} & \textbf{Method} & \textbf{Augmentation} & \textbf{Dist.} & \textbf{p10} \\
        \midrule
        \multirow{4}{*}{2024}  
        & KnightPupil    & \ding{55}  & --     & \textbf{96.66}  \\
        & KnightPupil    & \ding{51} & --     & 96.61  \\
        \cmidrule(lr){2-5}
        & Spatiotemporal~\cite{pei2024lightweightspatiotemporalnetworkonline}  & \ding{55}  & --     & 99.16 \\
        & Spatiotemporal~\cite{pei2024lightweightspatiotemporalnetworkonline}  & \ding{51} & --     & \textbf{99.37} \\
        \midrule
        \multirow{4}{*}{2025}
        & KnightPupil    & \ding{55}  & 2.82  & -- \\
        & KnightPupil    & \ding{51} & \textbf{2.78}  & -- \\
        \cmidrule(lr){2-5}
        & Spatiotemporal~\cite{pei2024lightweightspatiotemporalnetworkonline}  & \ding{55}  & 1.70  & -- \\
        & Spatiotemporal~\cite{pei2024lightweightspatiotemporalnetworkonline}  & \ding{51} & \textbf{1.61}  & -- \\
        \bottomrule
    \end{tabular}
    \caption{Comparison of Euclidean Distance and P10 Accuracy across different methods.}
    \label{tab:euclidean_comparison}
\end{table}

The results demonstrate the effectiveness of our augmentation techniques in improving model performance across different architectures.  

For the 2024 dataset, we observe a marginal improvement in p10 accuracy when augmentation is applied. The KnightPupil model experiences a slight decrease from 96.66\% to 96.61\%, suggesting that augmentation may introduce minor variability in predictions. However, for the Spatiotemporal model, augmentation provides a clear improvement, increasing p10 accuracy from 99.16\% to 99.37\%. This suggests that the Spatiotemporal model benefits more from augmentation, potentially due to its ability to leverage enhanced temporal patterns.  

For the 2025 dataset, we assess the Euclidean distance (Dist.), where lower values indicate better tracking accuracy. Augmentation consistently improves performance for both models. The KnightPupil model slightly reduces distance from 2.82 to 2.78, while the Spatiotemporal model benefits more significantly, reducing from 1.70 to 1.61. These improvements indicate that augmentation helps refine feature extraction and spatial-temporal representations, particularly in the Spatiotemporal model.  

Overall, our findings suggest that augmentation is crucial in refining tracking accuracy. The Spatiotemporal model benefits more from augmentation compared to KnightPupil, likely because its inherent architecture is better suited for handling variations in temporal patterns. This highlights the importance of carefully selecting augmentation strategies tailored to specific model architectures.

\paragraph{Data Augmentation Techniques.}
To evaluate the impact of different augmentation strategies, we conduct an ablation study by selectively removing specific augmentations and measuring their effect on model performance. The results are presented in~\cref{tab:ablation_augmentation_techniques}.  

\begin{table}[htbp]
    \centering
    \footnotesize
    \vspace{0.2cm}
    \begin{tabular}{lcc}
    \toprule
    \multirow{2}{*}{\textbf{Augmentation}} & \multicolumn{2}{c}{\textbf{Dist.}} \\  
    \cmidrule(lr){2-3}
    & KnightPupil & Spatiotemporal~\cite{pei2024lightweightspatiotemporalnetworkonline} \\ 
    \midrule
    w/o Temporal Shift  & 3.28 & 1.67 \\
    w/o Spatial Flip    & 3.26 & 1.64 \\
    w/o Event Deletion  & 3.34 & 1.66 \\
    Full Augmentation   & \textbf{3.08} & \textbf{1.61} \\
    \bottomrule
    \end{tabular}
    \caption{Ablation study on the impact of different augmentation techniques. We report the Euclidean distance (Dist.) for each setting. KnightPupil is trained for 200 epochs instead of the full 600-epoch setting.}
    \label{tab:ablation_augmentation_techniques}
\end{table}

Removing \textbf{temporal shift} leads to a noticeable increase in Euclidean distance, dropping performance from 3.08 to 3.28 for KnightPupil and from 1.61 to 1.67 for the Spatiotemporal model. This highlights the importance of temporal alignment in event-based eye tracking.  

Similarly, disabling \textbf{spatial flip} slightly degrades performance, increasing the Euclidean distance to 3.26 for KnightPupil and 1.64 for the Spatiotemporal model. This suggests that spatial variations contribute to model robustness, though their impact is less pronounced than temporal shift.  

When \textbf{event deletion} is removed, performance also deteriorates (3.34 for KnightPupil and 1.66 for the Spatiotemporal model), indicating that selectively removing events helps reduce noise and improve generalization.  

Overall, the \textbf{full augmentation} strategy achieves the best performance, reducing the Euclidean distance to 3.08 for KnightPupil and 1.61 for the Spatiotemporal model. These results emphasize the effectiveness of our augmentation pipeline in enhancing tracking accuracy.
\section{Conclusion}
\label{sec:conclusion}

In this work, we enhance the Lightweight Spatiotemporal Network via temporal shift, spatial flip, and event deletion to mitigate real-world noise, while KnightPupil combines EfficientNet-B3 spatial encoding, bidirectional GRU temporal modeling, and LTV-SSM adaptive dynamics to address sparse event patterns. Together, these strategies balance deployable efficiency with algorithmic flexibility for neuromorphic vision systems. We hope future work will explore unifying these approaches and investigating asynchronous processing that harmonizes with the sparse nature of event cameras. Our efforts aim to contribute to the evolving dialogue on temporal modeling for neuromorphic systems, striving to balance real-world applicability with algorithmic creativity.
\raggedbottom
\pagebreak
{
    \small
    \bibliographystyle{ieeenat_fullname}
    \bibliography{main}
}


\end{document}